\documentclass[10pt,journal]{IEEEtran}
\usepackage{bbm}
\usepackage{amsmath,amssymb,amsfonts}
\usepackage{graphicx}
\usepackage{textcomp}
\def\BibTeX{{\rm B\kern-.05em{\sc i\kern-.025em b}\kern-.08em
    T\kern-.1667em\lower.7ex\hbox{E}\kern-.125emX}}
    
\usepackage{fancyhdr}
\usepackage{soul,framed}
\usepackage{xspace}
\usepackage{array}
\usepackage{eqparbox}
\usepackage{url}
\usepackage{longtable}
\usepackage{lipsum}
\usepackage{blindtext}
\usepackage{makecell}
\usepackage{mathtools}
\usepackage{commath}
\usepackage{multirow}
\usepackage{tabularx}
\usepackage[normalem]{ulem}
\usepackage{xspace}
\usepackage{algorithm}
\usepackage{algpseudocode}
\usepackage{amsfonts}
\usepackage{placeins}
\usepackage{amssymb}
\usepackage{multirow}
\usepackage{tabularx}
\usepackage{caption}
\usepackage{subcaption}
\usepackage{pifont}

\usepackage{xspace}
\usepackage{algorithm}
\usepackage{algpseudocode}
\usepackage{amsfonts}
\usepackage{placeins}
\usepackage{booktabs}
\usepackage[dvipsnames]{xcolor}
\def\BibTeX{{\rm B\kern-.05em{\sc i\kern-.025em b}\kern-.08em
    T\kern-.1667em\lower.7ex\hbox{E}\kern-.125emX}}

\usepackage[pagebackref=true,breaklinks=true,colorlinks,bookmarks=false]{hyperref}
\hypersetup{
    allcolors=black
}

\definecolor{darkgreen}{RGB}{119,185,0}

\ifCLASSOPTIONcompsoc
  \usepackage[nocompress]{cite}
\else
  \usepackage{cite}
\fi

\begin{document}
\title{In-Distribution and Out-of-Distribution Self-supervised ECG Representation Learning for Arrhythmia Detection}
\author{Sahar Soltanieh, Javad Hashemi, and Ali Etemad
\thanks{This work was supported in part by Natural Sciences and Engineering Research Council of Canada.}
\thanks{Sahar Soltanieh is with the Department of Electrical and Computer Engineering, Queen's University, Kingston, ON K7L 3N6, Canada (e-mail: sahar.soltanieh@queensu.ca).}
\thanks{Javad Hashemi is at the School of Computing, Queen's University, Kingston, ON K7L 3N6, Canada (e-mail: javad.hashemi@queensu.ca).}
\thanks{Ali Etemad is with the Department of Electrical and Computer Engineering, \& Ingenuity Labs Research Institute, Queen's University, Kingston, ON K7L 3N6, Canada (e-mail: ali.etemad@queensu.ca).}}

\maketitle

\begin{abstract}
This paper presents a systematic investigation into the effectiveness of Self-Supervised Learning (SSL) methods for Electrocardiogram (ECG) arrhythmia detection. 
We begin by conducting a novel analysis of the data distributions on three popular ECG-based arrhythmia datasets: PTB-XL, Chapman, and Ribeiro. To the best of our knowledge, our study is the first to quantitatively explore and characterize these distributions in the area.
We then perform a comprehensive set of experiments using different augmentations and parameters to evaluate the effectiveness of various SSL methods, namely SimCRL, BYOL, and SwAV, for ECG representation learning, where we observe the best performance achieved by SwAV. Furthermore, our analysis shows that SSL methods achieve highly competitive results to those achieved by supervised state-of-the-art methods. To further assess the performance of these methods on both In-Distribution (ID) and Out-of-Distribution (OOD) ECG data, we conduct cross-dataset training and testing experiments. Our comprehensive experiments show almost identical results when comparing ID and OOD schemes, indicating that SSL techniques can learn highly effective representations that generalize well across different OOD datasets. This finding can have major implications for ECG-based arrhythmia detection. Lastly, to further analyze our results, we perform detailed per-disease studies on the performance of the SSL methods on the three datasets. 
\end{abstract}


\section{Introduction}
\label{sec:introduction}
\IEEEPARstart{T}he heart is a vital organ in the human body, and its study is of high importance. 
Electrocardiogram (ECG) is the most commonly used diagnostic technique for detecting and quantifying the electrical activity of the heart
\cite{1}. ECG analysis can help identify a range of heart conditions including, arrhythmia, heart attacks, and coronary artery disease. Traditional ECG analysis methods rely on the manual interpretation of trained experts, which can be labor-intensive and susceptible to inter-observer variability. 
In recent years, deep learning has emerged as a promising approach for the diagnosis of heart conditions based on ECG signals and has shown encouraging results in terms of performance and efficiency.
\textit{Supervised learning} is a popular paradigm in deep learning which relies on learning from input data along with respective output labels. However, the reliance on output labels for training the model can be viewed as a limitation, as the model can only learn to recognize classes of data that are included in the training dataset, and it may not be able to extract deeper features that are not directly related to the labels. Moreover, labelling medical datasets is a challenging task as it requires specialized experts with a high level of knowledge in the relevant medical field. This can make the process costly, time-consuming, and prone to errors or inconsistencies. 

However, recent advances in Self-Supervised Learning (SSL) have begun to overcome some of these difficulties. SSL uses unlabeled data to train a model, allowing it to learn from the data itself without the need for labels. This can make the training process more efficient and effective, and it can also enable the model to extract deeper representations which could lead to more accurate predictions. Self-supervision has shown great promise in a wide range of applications, including computer vision \cite{chen2020simple}, and natural language processing \cite{lan2019albert}.

While deep learning models have shown great potential in various fields, they are often trained and tested on similar data which contain similar distributions. This can lead to unexpected behaviour and safety hazards when models are presented with different data that come from different distributions. Out-of-Distribution (OOD) training and testing is a promising approach to address this issue by explicitly evaluating models on OOD inputs, for instance by evaluating models on datasets that are significantly different from the training data (e.g., recorded in different conditions, using different sensors, etc.). 
A number of recent studies \cite{ren2019likelihood, liu2020energy, ahuja2021invariance} have explored this notion in depth and made significant contributions to the area.

In this paper, we conduct a comprehensive analysis of SSL techniques for arrhythmia classification from ECG signals. Our study evaluates the performance of popular self-supervised methods across diverse datasets and assesses their generalizability and OOD classification capabilities. The results demonstrate that SSL achieves high accuracy in arrhythmia classification and enhances the performance of ECG-based diagnostic systems. Furthermore, we explore the potential of cross-dataset training to understand the adaptability of these models to different data distributions. Our findings provide valuable insights into the capabilities and limitations of SSL in arrhythmia ECG analysis, opening up promising avenues for future research.

In this research, we first analyze the distribution of samples using three arrhythmia-based ECG signal datasets, namely PTB-XL, Chapman, and Ribeiro. Based on the results of this analysis, we conclude that these datasets are suitable for our cross-dataset analysis and OOD experiments due to their different distributions. Accordingly, we select PTB-XL and Chapman datasets as pre-training datasets since they have a large number of samples, and select PTB-XL, Chapman, and Ribeiro test sets for testing the models.

Next, we select three popular and effective SSL methods, namely \textcolor{black}{Simple Framework for Contrastive Learning of Visual Representations (SimCLR)}, which is a highly popular contrastive learning method, \textcolor{black}{Bootstrap Your Own Latent (BYOL)}, which is a strong method that doesn't use any negative pairs, and \textcolor{black}{Swapping Assignments Between Views (SwAV)}, which utilizes a cluster code book to categorize the features in the datasets. We select these methods for their strong performances through the use of different self-supervision approaches.
To train these SSL methods, 
we select several commonly used augmentations for the ECG signals. 
Moreover, we choose multiple parameter sets for each augmentation based on prior research in the field. 
Finally, we pre-train each method in ID and OOD settings, followed by linear evaluation and finetuning.

Our extensive experiments reveal that SwAV consistently achieves the best overall results for ECG arrhythmia detection across all explored datasets in both ID and OOD settings.
Moreover, our cross-dataset analysis demonstrates that SSL methods, regardless of the technique, generally achieve comparable performances for ID and OOD data, indicating that SSL effectively enhances model generalization to OOD scenarios. 
Finally, our per-class analysis highlights the significant influence of per-class distributions on the classification performance of the models.

Our contributions in this work are summarized as follows: 
\begin{itemize}
\item We adapt and implement various SSL techniques and evaluate their effectiveness through a systematic study of various hyper-parameters. To achieve this goal, we adopt and implement the following popular SSL methods: SimCLR, BYOL, and SwAV. By carefully studying multiple augmentations and hyper-parameters, we present the efficacy of these techniques in the context of arrhythmia detection from ECG signals.
    
\item We assess the generalizability of different approaches for handling OOD and ID data. To evaluate the strengths and weaknesses of each approach, we use multiple datasets and experimental setups and evaluate the performance of the aforementioned SSL methods across ID and OOD training-testing schemes.
    
\item We perform a detailed disease-specific investigation on the effects of various SSL methods and associated parameters on different classes of heart disease. We explore the effects of different augmentations and their parameters on various types of arrhythmias. 
\end{itemize}

\section{Background and Literature Review}
SSL has gained significant attention in the machine learning community in recent years. 
Various self-supervised approaches have been proposed and evaluated successfully in different application areas. These areas include image classification~\cite{chen2020simple}, natural language processing~\cite{lan2019albert}, and speech recognition~\cite{baevski2020wav2vec}. Moreover, SSL has shown promising results when applied to the analysis of ECG signals for tasks such as emotion recognition~\cite{sarkar2020self, sarkar2020self1, sarkar2021detection} and arrhythmia detection~\cite{kiyasseh2021crocs, soltanieh2022analysis}.

The study~\cite{cheng2020subject} demonstrated the effectiveness of contrastive loss for SSL of bio-signals by developing augmentation techniques and addressing inter-subject variability through subject specific distributions. Their results showed that promoting subject-invariance improved classification performance and yielded effective weight initialization, highlighting the importance of subject awareness in bio-signal representation learning. The paper~\cite{lan2022intra} introduced Intra-Inter Subject SSL, designed to improve the diagnosis of cardiac arrhythmias from unlabeled multivariate cardiac signals. This method captured the temporal dependencies between heartbeats through intra- and inter-subject procedures.

The 3KG method introduced in \cite{gopal20213kg}, involves projecting 12-lead signals onto a 3D space and applying augmentations in that space. This approach has demonstrated strong results when fine-tuned for various heart diseases. Another approach, CLOCS, presented in \cite{kiyasseh2021clocs} utilizes patient-specified characteristics to achieve state-of-the-art performance in learning spatiotemporal representations of ECG signals in a contrastive learning manner. Lastly, \cite{oh2022lead} presents a novel SSL method for pre-training ECG signals that incorporates both local and global contextual information. This method features a new technique called Random Lead Masking, which enhances the model's robustness to arbitrary ECG leads.
 
In our recently published paper~\cite{soltanieh2022analysis}, we conducted a systematic investigation of the effectiveness of various augmentations and their corresponding parameters in arrhythmia detection. Through our analysis, we identified the optimal ranges of complexities for different augmentations. Our findings revealed the crucial role of type of augmentations and their hyperparameters in successful training with ECG.

In~\cite{kiyassehpcps}, a method was proposed for learning representations of the cardiac state of a patient using a combination of contrastive and supervised learning called Patient Cardiac Prototype (PCP). The authors showed that these PCPs could be used to identify similar patients across different datasets and that the representations maintained strong generalization performance when used to train a network. An SSL method was proposed in~\cite{luo2021segment}, referred to as Segment Origin Prediction, for classifying arrhythmia from ECG. The proposed method utilized a technique of assigning labels to samples based on their origin without the need for manual annotations.

In~\cite{zhou2022contrastive}, a deep learning approach for reducing false arrhythmia alarms in intensive care units using CNNs to learn representations of physiological waveforms automatically was presented. The proposed method employed a contrastive learning framework with a Siamese network and a similarity loss from pair-wise comparisons of waveform segments over time. The approach was augmented with learned embeddings from a rule-based method to leverage domain knowledge for each alarm type. The paper~\cite{mehari2022self} presented a comprehensive assessment of self-supervised representation learning for clinical 12-lead ECG data. Finally, the authors adapted state-of-the-art self-supervised methods to the ECG domain and evaluated their quality based on linear evaluation performance and downstream classification task performance.

The authors of paper~\cite{lee2021self} suggested an SSL algorithm that utilized ECG delineation as a method for classifying arrhythmia. They demonstrated the effectiveness of this approach and also the algorithm's ability to transfer the features learned during pre-training on one dataset to a different dataset, resulting in improved performance. A group of customized masked autoencoders for SSL ECG representation learning called MAE family of ECG was presented in paper~\cite{zhang2022maefe}. The approach contained three customized masking modes that focused on different temporal and spatial features of ECG. The encoder was utilized as a classifier in downstream tasks for arrhythmia classification.

Our literature review reveals promising results in using SSL techniques for ECG representation learning, showcasing competitive performance to or surpassing fully supervised methods. However, existing works often suffer from limitations such as small training datasets affecting model generalizability, insufficient exploration of robustness across diverse data setups, and challenges in hyperparameter and augmentation selection. To overcome these limitations, our work addresses the critical need to comprehensively evaluate SSL techniques for ECG-based arrhythmia detection by adapting popular SSL methods for ECG learning, studying diverse hyper-parameters and augmentations, and most importantly, systematically assessing the generalizability of SSL models across different datasets and OOD settings. This is crucial for real-world applications of automated ECG-based arrhythmia classification and robust deployment in clinical settings.

\section{Methodology and Experiment Setup}
As discussed earlier, our goal in this paper is to evaluate the effectiveness of various SSL methods in ECG arrhythmia detection. In this section, we provide a comprehensive overview of the SSL methods used for this goal. Specifically, we select the SSL methods to cover a diverse range of approaches. The first method is SimCLR, a popular SSL approach that utilizes contrastive learning. The second method used in this paper is BYOL, which belongs to the SSL category that exclusively operates on positive pairs without the use of negative pairs. Finally, we use Swav, which is known for its ability to use a cluster code book and feature clustering. The following section also offers an in-depth description of the datasets used in our experiments, followed by implementation details and evaluation protocols. We also provide a detailed account of the implemented augmentations. Lastly, we describe the protocol used for calculating the data distribution.

\subsection{Self-supervised Learning Techniques}\label{app:ssl}
In this section, we present the details of the SSL techniques that we use in this study, including, SimCLR, BYOL, and SwAV. 

\subsubsection{SimCLR}
\begin{figure}[!t]
    \centerline{\includegraphics[width=0.95\columnwidth]{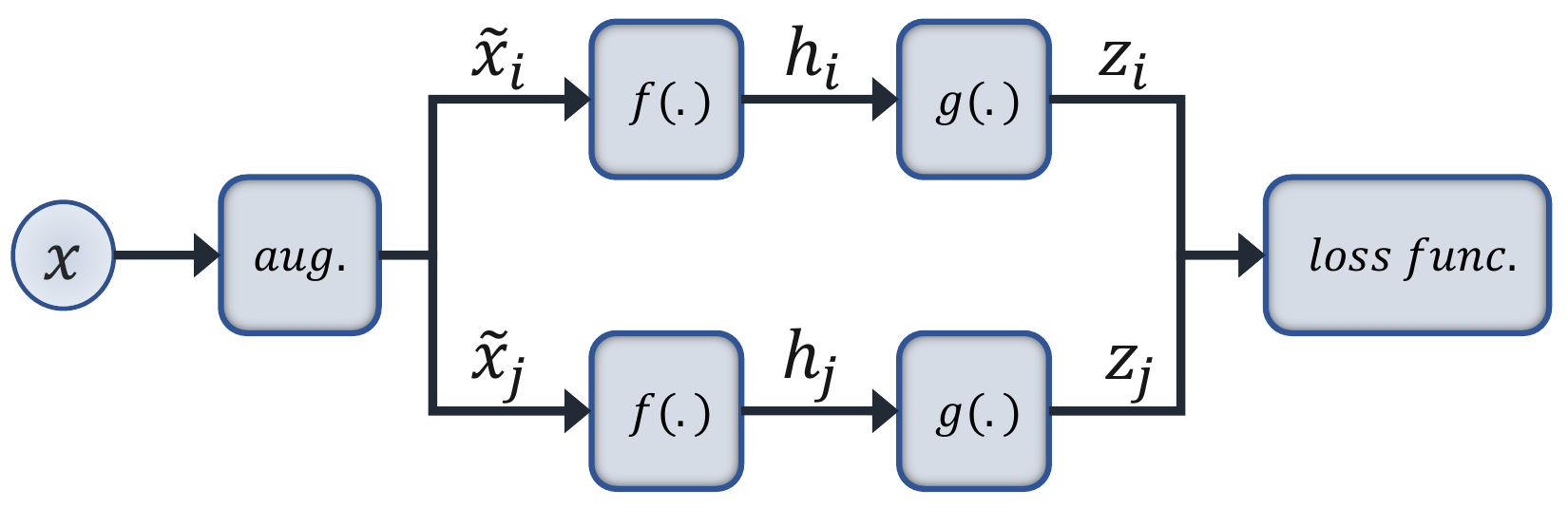}}
\caption{An overview of SimCLR.}
\label{fig:smiclr}
\end{figure}

The SimCLR~\cite{chen2020simple} method is a popular SSL technique that trains deep neural networks through contrastive learning. Contrastive learning is the notion of training a deep learning model by encouraging it to differentiate between pairs of similar and dissimilar data samples. In contrastive learning, the model learns to map samples from the same class closer together in a high-dimensional feature space, while pushing samples from different classes apart. The ultimate goal is to learn representations that capture meaningful features and patterns from the data, which can then be used for downstream tasks such as classification, segmentation, or clustering.
Figure~\ref{fig:smiclr} illustrates an overview of the SimCLR method. In this method, sample $x$ from the ECG dataset undergoes augmentations to generate two augmented signals, $\tilde{x_i}$ and $\tilde{x_j}$, which are two correlated versions of the original signal. The specific augmentations used are described in the following sections of the paper. As these two signals are derived from the same source signal, they are considered positive pairs and are treated as negative pairs with respect to all other samples. This approach allows SimCLR to learn without the use of class labels. Both augmented signals are then input into the encoder model $f(.)$, resulting in the final representations $h_i$ and $h_j$. These representations are then passed through the projection head $g(.)$ and the contrastive loss is calculated.

The \textcolor{black}{Normalized Temperature-Scaled Cross-Entropy (NT-Xent)} loss function introduced in \cite{sohn2016improved} aims to maximize the similarity between positive pairs while minimizing the similarity between negative pairs. As a result, it has been widely used in the SimCLR setup (which we also use in our study). The loss function is defined as follows:  
\begin{equation}
    \mathcal{L}_{i,j} = -\log\big(\frac
    {\exp(\mathrm{sim}(z_i, z_j)/\tau)}
    {\sum_{k=1} ^{2N} \mathbbm{1}_{(k\ne i)}\exp(\mathrm{sim}(z_i,z_k)/\tau)}\big).
\end{equation}

Here, for a batch of $N$ samples, by applying two augmentations per sample, the total data used for loss calculation becomes $2N$. The pairs $(z_i, z_j)$ are considered positive pairs, while they are considered negative pairs with respect to the remaining $2N-2$ samples. The temperature parameter $\tau$ serves to adjust the slope of the loss function. Specifically, higher values of $\tau$ result in a smoother loss function, while lower values yield a steeper loss function. The indicator function $\mathbbm{1}$ in the denominator of the equation takes a value of 1 when the samples are negative pairs, and 0 when they are positive pairs. The similarity function $\mathrm{sim}$ employed in this equation is the cosine similarity, defined as follows:
\begin{equation}
    \mathrm{sim}(z_i, z_j)=\frac{\langle z_i, z_j \rangle}{\parallel z_i\parallel \parallel z_j\parallel} .
\end{equation}
Upon training the model using the contrastive learning paradigm, the encoder model, $f(.)$, is used as a pre-trained feature extractor which could be used as frozen or fine-tuned for better domain alignment.

\subsubsection{BYOL} 
\begin{figure}[!t]
    \centerline{\includegraphics[width=0.95\columnwidth]{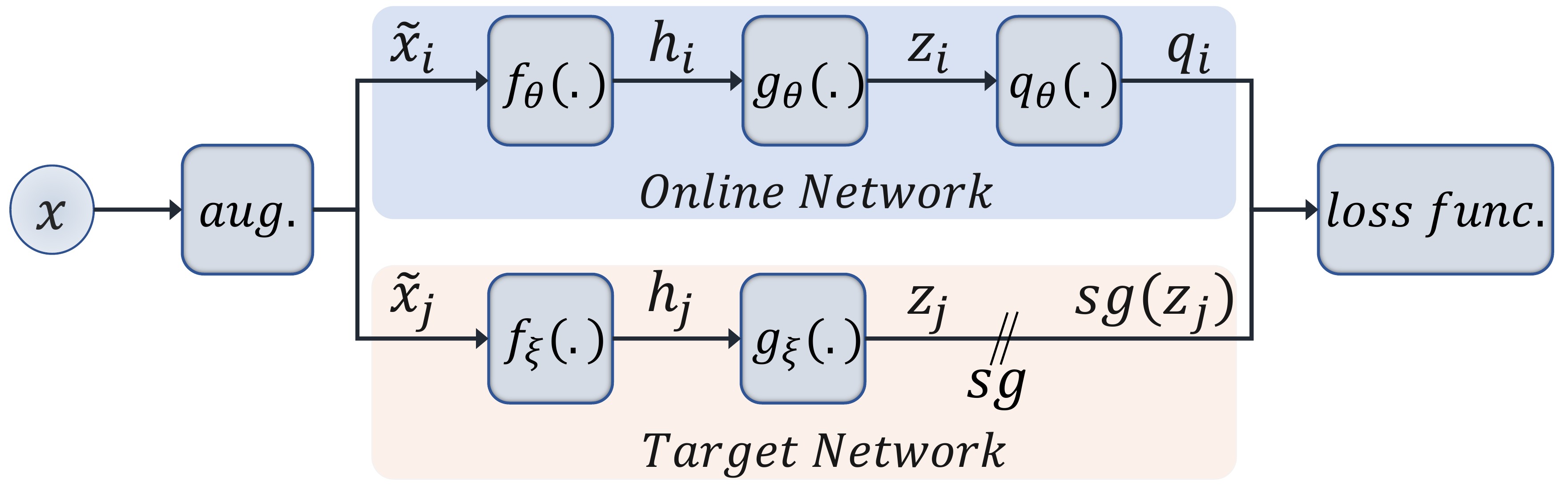}} 
\caption{An overview of BYOL.}
\label{fig:byol}
\end{figure}

BYOL~\cite{grill2020bootstrap} is an SSL technique that uses positive pairs to train deep neural networks. A visual representation of this method can be seen in Figure~\ref{fig:byol}. This method involves the use of two networks, referred to as the \textit{online} and \textit{target} networks, which learn from each other by comparing the representation of an input sample as it passes through each network. Specifically, the \textit{online} network is trained to recognize that the sample passing through the \textit{target} network is the same as the original input sample. This interaction between the two networks allows for effective learning without the need for explicit labels.

In this method, a sample $x$ is first processed through an augmentation block, resulting in two different augmented versions, $\tilde{x_i}$ and $\tilde{x_j}$, which represent distinct perspectives of the same input sample. The first augmented version, $\tilde{x_i}$, is then passed through the \textit{online} network encoder, denoted as $f_\theta(.)$, which produces a deeper representation $h_i$ of the sample. This representation is further processed through the \textit{online} projection head, $g_\theta (.)$, to produce $z_i$. From the other side of the network, the second augmented version of the sample $x$, $\tilde{x_j}$, is passed through the \textit{target} encoder, $f_\zeta (.)$, and projection head, $g_\zeta (.)$, to produce the representations $h_j$ and $z_j$, respectively. 

Since the \textit{online} and \textit{target} networks are both trained using different versions of the same sample, they are anticipated to contain similar information. The \textit{online} network is trained to predict the \textit{target} network representation, thus eliminating the need for negative pair training. This is achieved through the use of the function $q_\theta (.)$, which serves as a prediction mechanism for the \textit{online} network to match the representations generated by the \textit{target} network. The final representation of the \textit{online} network, $z_i$, is passed through the predictor function $q_\theta (.)$, resulting in $q_i$. Meanwhile, a stop gradient (denoted as $sg$) is applied to the final representation of the \textit{target} function, $z_j$, as only the weights of the \textit{online} network will be updated at this stage. By learning to predict the representations generated by the \textit{target} network, the \textit{online} network encodes the underlying features of the input data effectively, leading to robust and highly representative feature extraction.

The loss function is designed to encourage the outputs of the \textit{online} and \textit{target} networks, represented by $q_i$ and $z_j$, to become more similar. This is achieved by calculating the mean squared error between the L2-normalized output of the \textit{online} network's predictor, and the final representation of the \textit{target} network. The resulting loss is expressed in the following: 
\begin{equation}
    \mathcal{L}_{\theta,\zeta} = \Vert\bar{q_i}-\bar{z_j}\Vert^2 _2 = 2-2.(\frac{\langle q_i,z_j \rangle}{\Vert q_i\Vert_2 \Vert z_j\Vert_2}),
\end{equation}
where $\bar{q_i}, \bar{z_j}$ are the L2-normalized form of $q_i, z_j$, and $\langle q_i,z_j \rangle$ is the dot product of $q_i, z_j$. 

To ensure symmetry in the loss calculation, the input $\tilde{x_i}$ is fed to the \textit{target} network, while $\tilde{x_j}$ is fed to the \textit{online} network. The resulting loss, denoted as $\tilde{\mathcal{L}}_{\theta,\zeta}$, is computed. The final loss is given by $\mathcal{L}^{BYOL}_{\theta,\zeta}=\mathcal{L}_{\theta,\zeta}+\tilde{\mathcal{L}}_{\theta,\zeta}$, which combines the loss calculated from both the \textit{online} and \textit{target} networks and their flipped versions.

In each iteration of the training process, the \textit{target} network's weights are updated by taking a weighted average of the \textit{online} network's weights and the \textit{target} network's previous weights. The weight applied in this average, known as the exponential moving average decay, is predetermined and serves to gradually incorporate the weights of the online network into the target network while maintaining some of the target network's own previous weights to stabilize training.

One of the key benefits of BYOL is its use of SSL without the need for negative pairs. This reduces the difficulties associated with selecting negative pairs, such as the challenge of finding pairs that are appropriately challenging, as well as the problem of curriculum learning. The predictor function, $q_\theta(.)$, is specifically designed to learn to identify the semantic segments that are important for ECG signal classification, while ignoring irrelevant information, such as noise or errors in signal acquisition. This results in a model that is more robust to hyper-parameters and able to accurately classify ECG signals.

\subsubsection{SwAV} 
\begin{figure}[!t]
    \centerline{\includegraphics[width=.95\columnwidth]{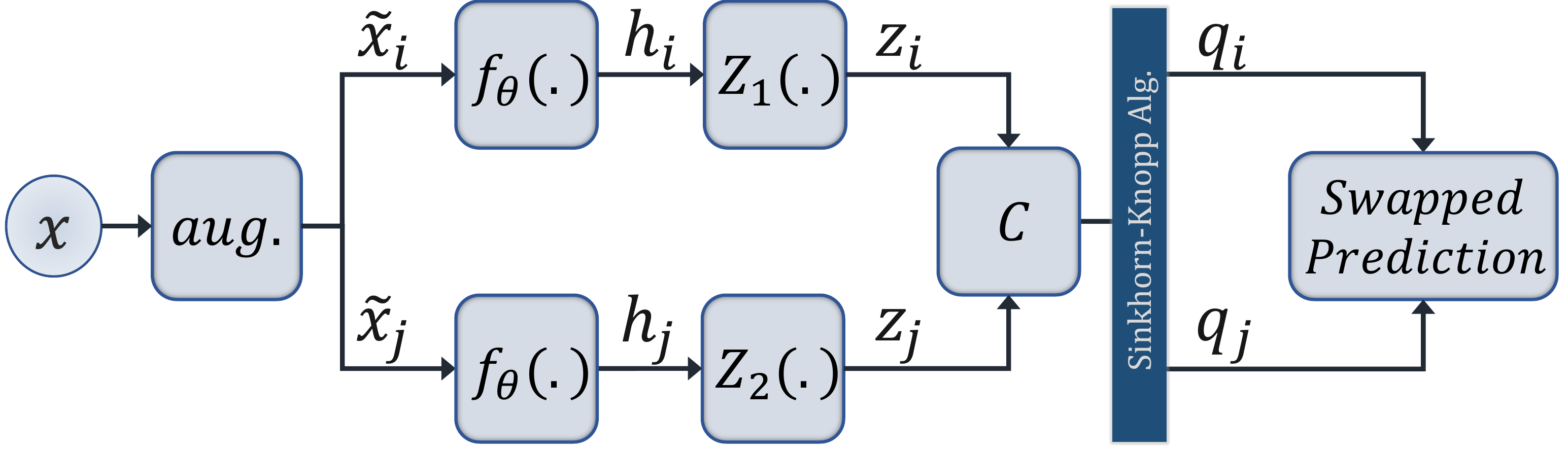}}   
\caption{An overview of SwAV.}
\label{fig:swav}
\end{figure}

SwAV~\cite{caron2020unsupervised} is an innovative clustering-based SSL approach that eliminates the need for pairwise comparisons, as briefly depicted in Figure~\ref{fig:swav}. Unlike the two previous methods, the SwAV approach slightly changes the way the loss function and augmentation are implemented. This method starts by obtaining representations of the input data after undergoing various transformations. These representations are then assigned to codes using a cluster codebook and the Sinkhorn-Knopp algorithm~\cite{cuturi2013sinkhorn}. The model compares the representations of each path with the codes assigned to the opposite path and attempts to minimize the loss function and bring the representations closer together. In the following paragraphs, we will provide a more in-depth explanation of the SwAV method.

This method begins by applying augmentations to the ECG signal $x$ to obtain different views of the input signal, denoted as $\tilde{x_i}$ and $\tilde{x_j}$. These views are then processed through the network $f_\theta (.)$, resulting in representations $h_i$ and $h_j$. These representations are passed through shallow non-linear networks, $Z_1(.)$ and $Z_2(.)$, to produce $z_i$ and $z_j$. The next step involves utilizing a cluster codebook, or a set of prototypes, denoted as $C=[c_1, c_2, ..., c_K]$, to assign representations to their corresponding codes. These prototype vectors are trained using the dominant features of the dataset. The prototype vectors can also be thought of as the weights of a single-layer fully connected network, where the dot product between the weights and the representations is calculated. The Sinkhorn-Knopp algorithm~\cite{cuturi2013sinkhorn} is then applied to generate the codes for each sample, $q_i$ and $q_j$.

The next step in the process involves computing the loss for weight updates using the swapped prediction method. This involves utilizing the codes of sample $q_i$ to predict the representation of sample $z_j$, and vice versa. The loss is calculated by computing the cross entropy between the codes and the softmax of the dot product of the representations and cluster codes. This calculation enables the optimization of the weights, leading to improved predictions. The loss function for this process is given by the following equations:
\begin{equation}
    \mathcal{L}_{z_i,z_j} = l(z_i, q_j)+l(z_j, q_i),
\end{equation}
\begin{equation}
 l(z_a, q_b) = - \sum_{k} {q_b^{k} \log \frac{\exp(\frac{1}{\tau} z_a^{T} C_k)}{\sum_{k^{'}} {\exp(\frac{1}{\tau}z_a^T {C_{k^{'}}})}}},
\end{equation}
where the temperature parameter, $\tau$, is used to regulate the confidence or uncertainty of the predictions. The higher the temperature value, the softer or more uncertain the output will be, and the lower the temperature value, the harder or more confident the output will be. Upon completion of training, the network $f_\theta(.)$ can be detached from the framework and be used as a powerful feature extractor for ECG signals or other types of data.

\subsection{Datasets}\label{app:dataset}

\begin{table}[!t]
  \centering
  \caption{Summary of datasets used in this paper.}
  \label{tab:dataset-summary}
  \begin{tabular}{lccc}
  \toprule
     Dataset &  \# of Subjects &  \# of Samples &  Reference\\
    \midrule
    PTB-XL & 18,885 & 21,837 & \cite{wagner2020ptb}  \\
    Chapman & 10,646 & 10,646 & \cite{zheng202012} \\
    Ribeiro & 827$^*$ & 827$^*$ & \cite{ribeiro2020automatic}\\
    \bottomrule
  \end{tabular}
  
  \vspace{5pt} 
 \textit{* This is the number of patients and samples in Ribeiro dataset designated by the authors of \cite{ribeiro2020automatic} as the test set.}
\end{table}

In this study, we aim to evaluate the performance of the aforementioned SSL methods in the context of ID and OOD data. To this end, we use several popular ECG-arrhythmia datasets, namely, PTB-XL~\cite{wagner2020ptb}, Chapman~\cite{zheng202012}, and Ribeiro~\cite{ribeiro2020automatic}. Below we provide the characteristics and details of these datasets, and Table~\ref{tab:dataset-summary} provides a summary of the datasets used in this study.

\noindent \textbf{PTB-XL.} PTB-XL~\cite{wagner2020ptb} dataset is a publicly available resource and one of the largest of its kind for ECG signals. It serves as a widely used benchmark for evaluating the performance of models on 12-lead ECG signals, consisting of $18,885$ patient records, totalling $21,837$ records of $10$-second duration each, collected using Schiller AG devices between 1989 and 1996. The dataset is available at two different sampling frequencies: $100Hz$ and $500Hz$. In this study, we use the $100Hz$ version to limit the required computational resources. PTB-XL dataset is comprised of three categories: Diagnostic, Form, and Rhythm, amounting to a total of 71 classes.
In this study, we focus on the Diagnostic labelling category, which contains $5$ main categories.
The dataset is relatively balanced in terms of gender, with $52\%$ male and $48\%$ female patients, and has a wide age range from $0$ to $95$ with a median of $62$ years.

\noindent \textbf{Chapman.}
The \textcolor{black}{Shaoxing} People's Hospital and Chapman University have jointly collected and published a dataset in $2020$ that contains 12-lead ECG signals from $10,646$ patients~\cite{zheng202012}. The dataset includes ECG signals with both arrhythmia and various cardiovascular diseases, including $11$ rhythm conditions.
The ECG signals are recorded with a duration of $10$ seconds at a sampling frequency of $500Hz$ and have undergone necessary pre-processing steps to ensure consistency with other datasets. This dataset is chosen for its size, comprehensiveness, and representation of various heart conditions. The age range of the participants in the dataset is from $4$ to $98$ years old, and it is relatively gender balanced with $44\%$ female and $56\%$ male participants. $17\%$ of the dataset consists of samples with a normal heartbeat, while the remaining $83\%$ have been diagnosed with a specific type of heart abnormality.

\noindent \textbf{Ribeiro.}
The Telehealth Network of Minas Gerais in Brazil has compiled a dataset (Ribeiro~\cite{ribeiro2020automatic}) of ECG signals for the purpose of evaluating the performance of algorithms for the diagnosis of cardiovascular abnormalities. The dataset includes samples from patients diagnosed with one of six specific abnormalities. The dataset is divided by the original authors of the dataset ~\cite{ribeiro2020automatic} into standard training and test sets.
The age of the patients in the test set ranges from $16$ to over $81$ years old, and the dataset exhibits a bias towards female patients, with $38.8\%$ of the patients being male and the rest being female. The sampling frequency is $400Hz$ which is pre-processed for the sake of consistency with other datasets. All of the conditions in Ribeiro dataset can fall under the main category of Conduction Disturbance (CD), specifically, 1st-degree AV block (1dAVb), Right Bundle Branch Block (RBBB), Left Bundle Branch Block (LBBB), Sinus Bradycardia (SB), Atrial Fibrillation (AFIB). 

\subsection{Implementation Details and Evaluation Protocol}\label{app:imp_detail}

We divide PTB-XL and Chapman into three distinct portions for training, validation, and testing. More specifically, in the case of PTB-XL, the dataset is divided into $10$ subsections recommended by the original authors of \cite{wagner2020ptb}, based on which we create the three sets using $8$ folds for training, $1$ fold for validation, and $1$ fold for testing. For Chapman, given a lack of standardized subsets, we randomly select $80\%$ of the samples for training, $10\%$ for validation, and the remaining $10\%$ for testing. Finally, Ribeiro is only used for fine-tuning/testing, unlike PTB-XL and Chapman which are commonly used for pre-training. Accordingly, we divide the test set of Ribeiro into $3$ distinct folds where $80\%$ of the test subjects are used fine-tuning, $10\%$ are used for validation, and $10\%$ are used for testing.

Next, we down-sample Chapman and Ribeiro datasets from $500Hz$ and $400Hz$ respectively to $100Hz$. PTB-XL dataset is available in two versions, $100Hz$ and $400Hz$, and we opt for the former.
Finally, as common in the area \cite{mehari2022self}, we window the signals into $2.5$-second segments with no overlap, resulting in samples consisting of $250$ data points each.

During the pretraining phase, we optimize the NT-Xent loss which is described in section~\ref{app:ssl}. In this phase, the best model is selected based on the lowest loss observed on the validation set. In the fine-tuning phase, we use a binary cross-entropy loss and select the best model based on the highest  f1 score observed on the validation set.

For training, we implement our method using PyTorch and utilize a pair of NVIDIA GeForce GTX 2080 GPUs.
To train both the contrastive framework and supervised baselines we utilize the Adam optimizer with a learning rate of $5\times10^{-4}$ and a weight decay of $1\times10^{-3}$. For fine-tuning, we reduce the learning rate to $5\times10^{-3}$ to prevent catastrophic forgetting and minimize the chance of overwriting the network weights completely. The contrastive phase is trained for $2000$ epochs with a batch size of $4096$ for SimCLR and BYOL, and a batch size of $1024$ for SwAV due to computational resource constraints. For the fine-tuning phase, we train the model for $50$ epochs. Parameters are carefully chosen to maximize performance and considering prior works, e.g., \cite{oh2022lead, mehari2022self, kiyasseh2021clocs}.

We utilize the xResNet1d50 model as our backbone model for the experiments, as it was shown in \cite{strodthoff2020deep}. This model is an adaptation of ResNet50 model for 1-dimensional data, \textcolor{black}{and is known to be an effective backbone that provides stable and consistent performance for ECG representation learning}.

To evaluate performance, we use the macro  f1 score instead of the micro  f1 score with a threshold of 0.5 as commonly used in this area (\cite{sarkar2020self, clifford2017af, christov2017multi}). This is because the micro  f1 score assigns equal weight to all observations, while the macro  f1 score assigns equal weight to each class. Since most medical datasets, including ECG datasets, are imbalanced across classes, it is more reasonable to use a performance metric that treats all classes equally. Additionally, many medical datasets have more negative samples than positive ones. Therefore, metrics such as accuracy, which treats all samples equally, may not be suitable. By using the macro  f1 score, we can accurately evaluate our model's performance on imbalanced datasets while ensuring that each class is given equal weight. 

\subsection{Augmentation Details}\label{app:aug} 
\noindent \textbf{1) Gaussian Noise.}
We add a noise signal $N(t)$ to ECG signal $x(t)$. $N(t)$ is the same size as the ECG sample and each of its data points is driven from a Gaussian distribution with a mean equal to $0$, and a standard deviation equal to the adjustable parameter.  
We examined $\sigma_{G}=[0.01, 0.1, 1]$ in our experiments.

\noindent \textbf{2) Channel Scaling.}
To obtain this augmentation, we multiply the ECG signal by a scaling factor $S$, which is a set of scaling factors for each of the 12 leads: $S = \{s_i \mid i = 1,2,\dots,12\}$. Each $s_i$ is randomly derived from the range $[a, b]$, where both $a$ and $b$ are selected as augmentation parameters. We tested scaling factors in the range of $S_{ch}=[(0.33, 3), (0.33, 1), (0.5, 2)]$ in our experiments.

\noindent \textbf{3) Negation.}
The input signal is subject to vertical flipping across the time axis, which can be expressed as $\tilde{x}(t) = -x(t)$.

\noindent \textbf{4) Baseline Wander.}
To simulate this type of noise in an ECG signal, we artificially introduce a low-frequency sinusoidal waveform with the frequency of $f_w$ to the original ECG signal. The sinusoidal waveform is generated with the chosen frequency, and the scale for the waveform is selected as $S_{bw}$. 
In our experiments, we choose the frequency of the sinusoidal $f_w = 100$ and we test scales $S_{bw}=[0.1, 0.7, 1]$.

\noindent \textbf{5) Electromyographic (EMG) Noise.}
To simulate EMG noise, high-frequency white Gaussian noise is used. This is because EMG noise is primarily caused by fast muscle contractions, which have high-frequency components~\cite{kher2019signal}. A Gaussian distribution with a mean of $0$ is used, and the standard deviation of the distribution can be adjusted to achieve the desired level of noise. 
We use a variance equal to: $\sigma_{EMG} = [0.01, 0.5, 1]$ in our experiments. 

\noindent \textbf{6) Masking.}
This augmentation technique involves zeroing out specific segments of each lead in the ECG signal. To apply this technique, we first select the higher and lower bound of windows that we want to set to zero. Accordingly, two percentage values $a$ and $b$ are chosen as this augmentation's parameters. Next, for each ECG signal in the batch, a random value $c$ is chosen from the range $[a, b]$, and using this value, 
a $c$\% segment of that lead is set to zero. In our experiments we examined the following ranges for the masking parameters: $[10\%, 20\%], [0\%, 50\%], [40\%, 50\%]$.

\noindent \textbf{7) Time Warping.}
We start this augmentation by dividing $x(t)$ into $w$ segments, denoted as $x_1(t)$, $\cdots$, $x_w(t)$. Next, we randomly select half of the segments and apply time warping to stretch them by a scaling factor of $r\%$ while simultaneously squeezing the other half by the same amount. Finally, we concatenate the segments in the original order to produce the augmented signal, denoted as $\tilde{x}(t)$.
We test the following list of parameters for this augmentation: $(w,r)=[(1, 10), (3, 5), (3, 10)]$.

\noindent \textbf{8) Combination of Augmentations.}
To further analyze the impact of augmentations on our pre-training process, we conduct additional experiments where we apply a combination of four augmentations simultaneously. This allows us to observe how different augmentations interact with each other and how this affects the performance of the models. In each iteration of pre-training, we randomly select four augmentations from our previously described list of augmentations. We select the parameters through experimentation with the goal of maximizing performance.
The parameters for the augmentations when combined are as follows: Gaussian noise ($\sigma_{G}=1$), channel scaling ($S_{ch}=(0.33, 3)$), baseline wander ($S_{bw}=1$), EMG noise ($\sigma_{EMG} = 0.01$), masking ($[40\%, 50\%]$), and time warping ($(w, r)=(1, 10)$).

\subsection{Data Distribution}\label{app:data_dist}
The distributions of data play a crucial role in this study, as we aim to investigate various SSL techniques in both ID and OOD settings. To achieve this goal, we thoroughly analyze the three datasets used in this study, namely Chapman, PTB-XL, and Ribeiro, with a focus on their data distributions. This analysis allows us to comprehend the relationships between these datasets and gain insights into their relative characteristics.

\begin{figure}[!t]
    \centerline{\includegraphics[width=0.88\columnwidth]{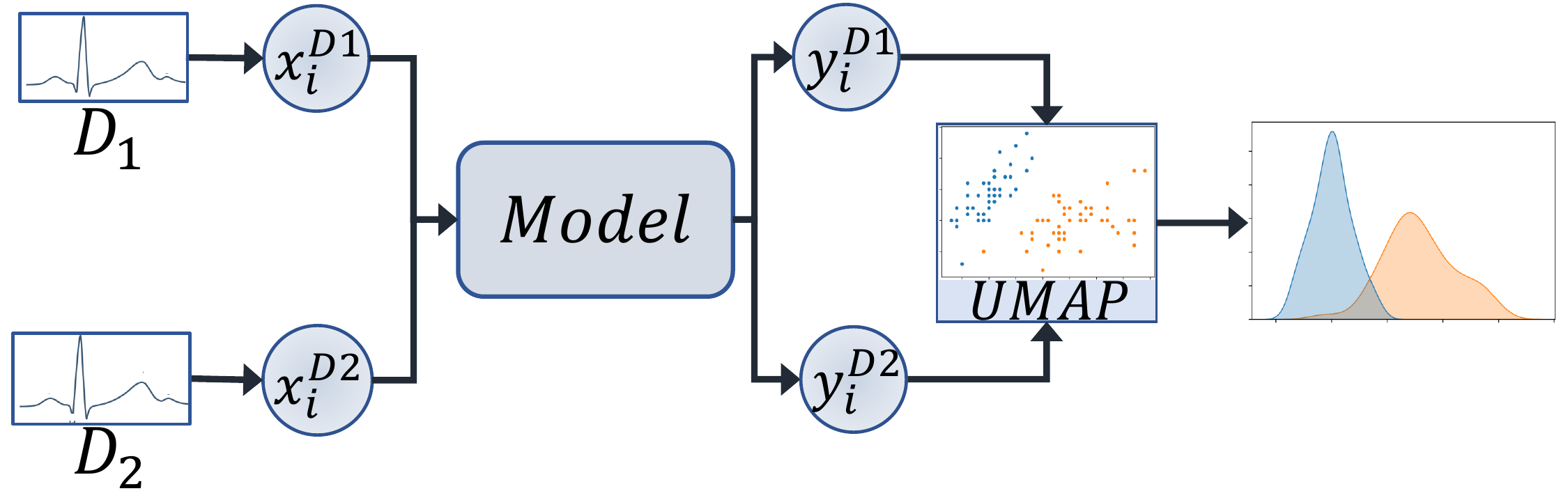}}
    \caption{Data distribution analysis framework.}
    \label{fig:data_distribution_framework}
\end{figure}

To perform this analysis, we use a pre-trained model to process every sample $x^{D_1}_{i}$ from a dataset ($D_1$) and produce their respective outputs $y^{D_1}_{i}$ as illustrated in Figure~\ref{fig:data_distribution_framework}. Similarly, samples $x^{D_2}_{i}$ from $D_2$ are passed through the same model to produce $y^{D_2}_{i}$. This approach of using the outputs to determine the dataset distributions has been previously employed in other studies as well~\cite{hendrycks2016baseline, wang2022vim, zhang2023decoupling}. Dimensionality reduction is then applied on the resulting outputs ($y^{D_1}_{i}$ and $y^{D_2}_{i}$) using the UMAP method \cite{mcinnes2018umap} with the number of neighbors equal to 30 and a minimum distance of 0.25. 
This analysis shows whether two datasets, $D_1$ and $D_2$, are ID or OOD based on the overlaps observed in their UMAP distributions. If UMAP distributions of the datasets have considerable overlap, the samples from $D_1$ and $D_2$ are considered ID. Conversely, if the distributions are weakly overlapped, they are considered OOD. 

To mathematically measure the overlap between the distributions of the two datasets, we adopt the method proposed in~\cite{pastore2019measuring}. After obtaining the lower-dimensional representations of $D_1$ and $D_2$ through UMAP ($\hat{y}^{D_1}_{i}$ and $\hat{y}^{D_2}_{i}$), we estimate their respective Probability Density Functions (PDFs).
To quantify the degree of overlap between the two distributions, we calculate the overlapping index ($\eta$) using the following equation:
\begin{equation}
    \eta(D_1, D_2) = \int_{-\infty}^{\infty} \min(f_X(x), f_Y(x)) \mathrm{d}x
\end{equation}
This formula calculates the area under the curve where the PDFs of the datasets overlap. A value of $\eta$ approaching $1$ indicates that the datasets are from similar distributions, whereas, if $\eta$ is significantly lower than $1$, it suggests that the datasets are OOD in relation to each other.

\section{Results and Discussions}\label{ch:5}
This section presents the results of our experiments and their implications. We begin by analyzing the distribution of the datasets used in this work, as described in Section \ref{app:data_dist}. Next, we evaluate the impact of different augmentation techniques on the performance of different SSL methods, as explained in Section \ref{app:aug}. This is followed by a comparison of the performance of the SSL methods themselves in the context of ID and OOD, as outlined in Section \ref{app:ssl}. Finally, we examine how different factors influence the model's ability to classify each disease class in the datasets.

\subsection{Data Distribution Analysis}\label{app:data_distribution_analysis}

\begin{figure}[!t]
    \centerline{\includegraphics[width=.9\columnwidth]{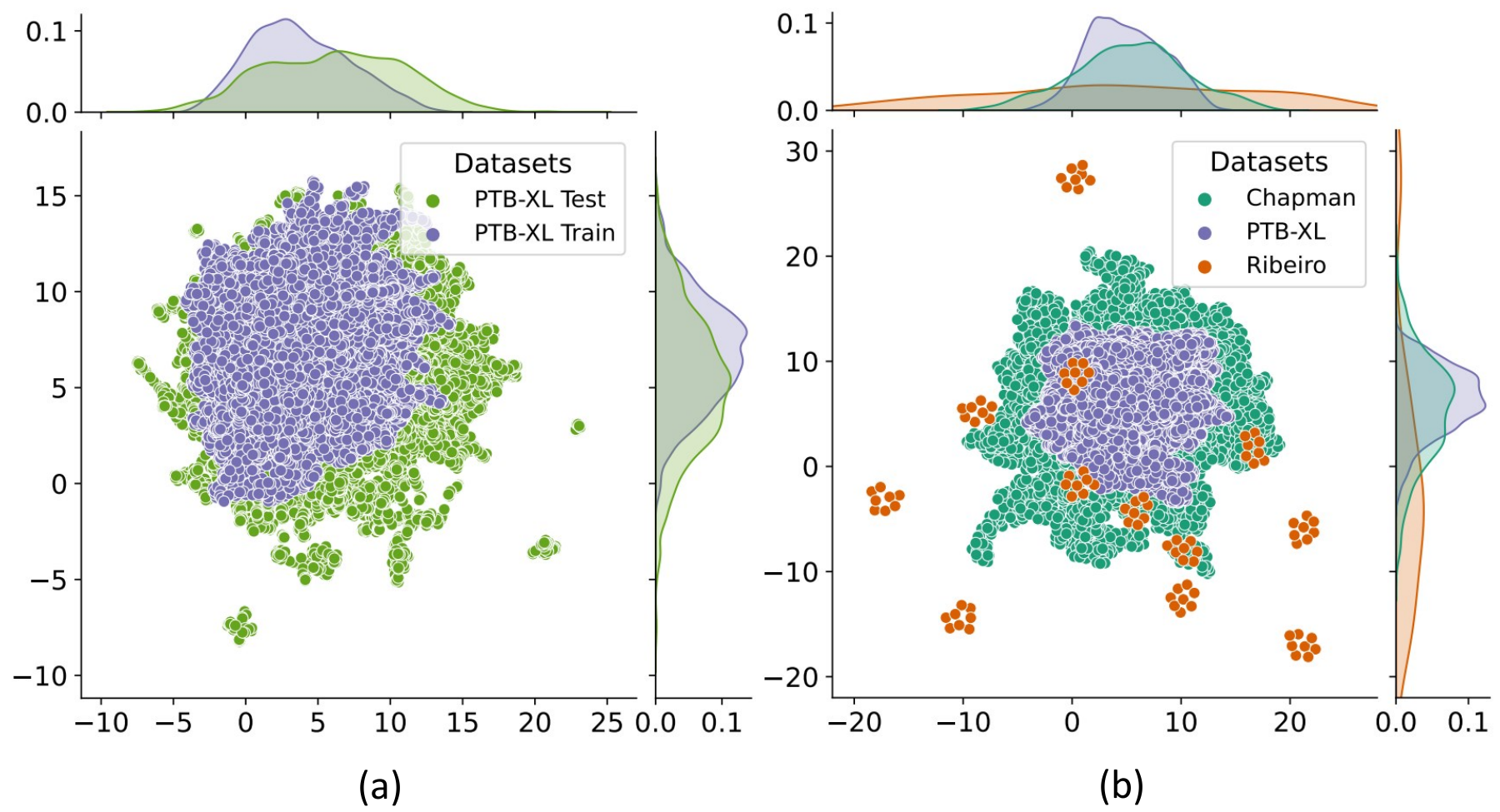}}
\caption{Visual representation of the distribution of the datasets. Part (a) illustrates the distributions of PTB-XL train and test subsets (ID) and part (b) shows the OOD instances (Chapman and Ribeiro) with respect to PTB-XL dataset.}
\label{fig:ptbxl_distributions}
\end{figure}

To demonstrate and analyze the distributions of the datasets used in this study, we use the training sets of the two larger datasets, PTB-XL and Chapman, for pre-training, while all three datasets Ribeiro, PTB-XL, and Chapman are used for distribution analysis.
For training the xResNet1d50 model from which to obtain the embeddings, the best-performing SSL method, SwAV, is used with time warping augmentation using parameters (3, 10).
Finally, dimensionality reduction is performed using UMAP. 
Figure~\ref{fig:ptbxl_distributions} (left) shows the outcome of this analysis for PTB-XL train and test sets, where we observe that the distributions expectedly fall within one another. In Figure~\ref{fig:ptbxl_distributions} (right) we present the embeddings of the other two datasets (Chapman and Ribeiro) when a model that is pre-trained on PTB-XL (training set) is used to obtain the embeddings. It can be seen that the embeddings of Chapman and Ribeiro are considerably outside the distribution of PTB-XL, making our configuration of data effective for cross-dataset OOD analysis.

\begin{figure}[!t]
    \centerline{\includegraphics[width=.9\columnwidth]{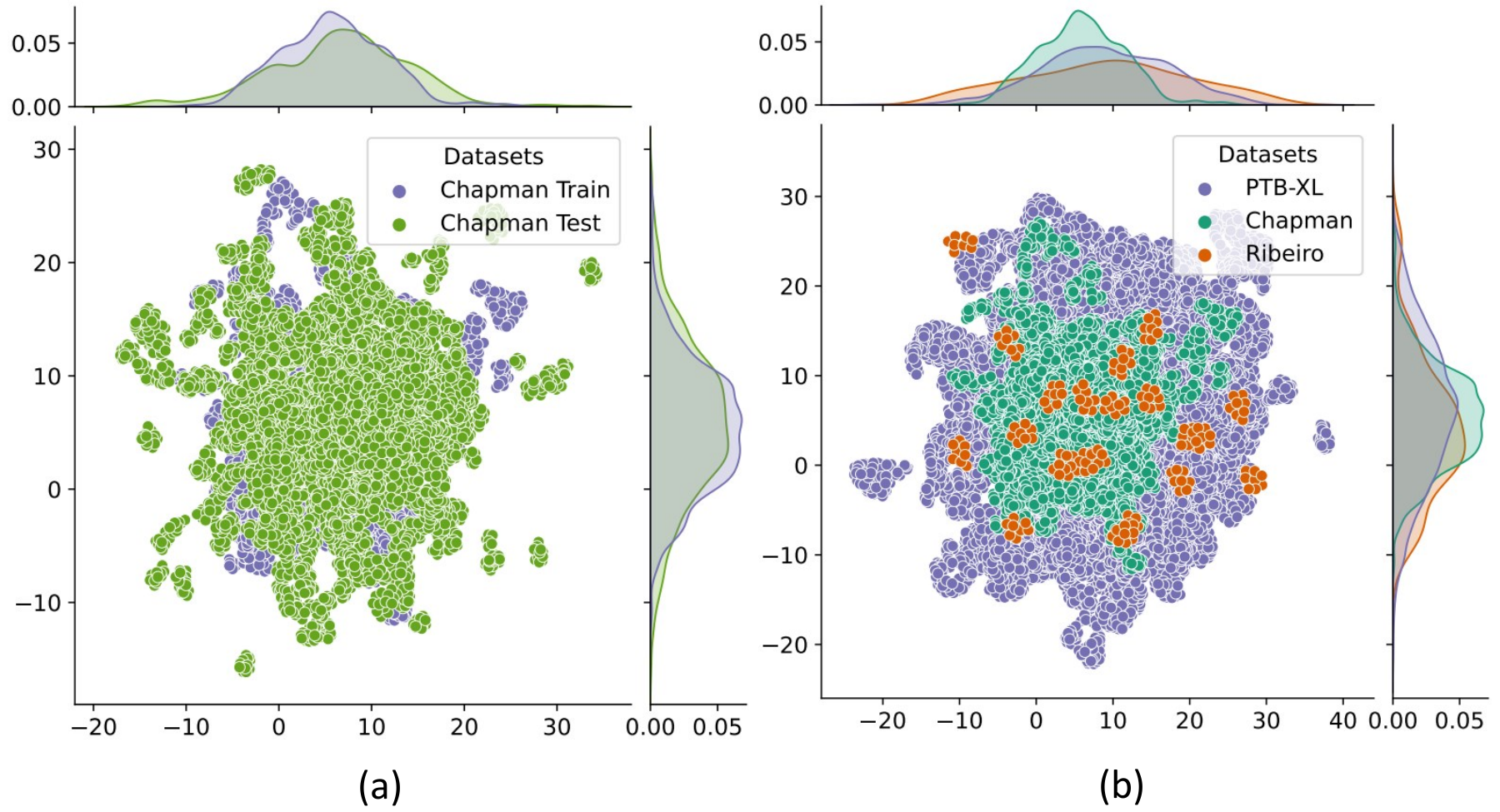}}
\caption{Visual representation of the distributions of the datasets used in our experiments. Part (a) illustrates the ID instances for Chapman dataset, while part (b) shows the OOD instances (PTB-XL and Ribeiro) with respect to Chapman dataset.}
\label{fig:chapman_distributions}
\end{figure}

\begin{figure}[!t]
    \centerline{\includegraphics[width=.95\columnwidth]{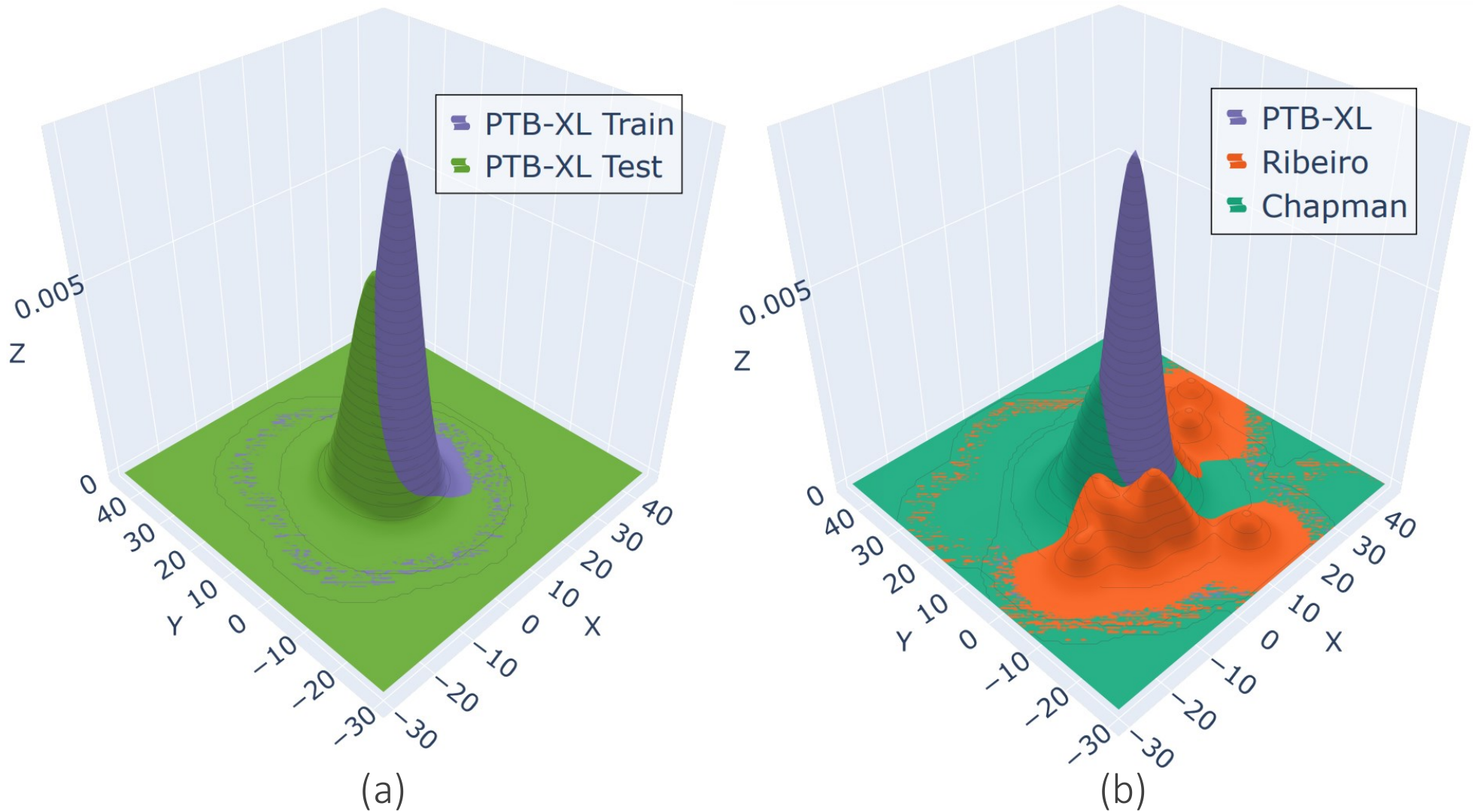}}
\caption{Comparing ID (PTB-XL train and test sets) against OOD (PTB-XL train set, Chapman, and Ribeiro) distribution overlaps through fitted KDE on the representations (left: ID, right: OOD).}
\label{fig:ptb_distributions_counts}
\end{figure}

Here, we aim to quantify the distribution shifts using the procedure outlined in Section \ref{app:data_dist}. This procedure involves applying a Kernel Density Estimation (KDE) on the UMAP outputs to estimate PDFs for the distributions. The overlap between the train and test sets of PTB-XL dataset is thus estimated by taking the integral of the overlapping area of the two KDEs. Our results, which are presented in Table~\ref{tab:PTB_distances} indicate that the overlap between the two distributions (PTB-XL train and test sets) is $83.52\%$. The same procedure is then applied to Chapman and Ribeiro datasets, to calculate the overlap between the train set of PTB-XL dataset and the other two datasets. The results show that the overlap between PTB-XL (train) and Chapman is $63.65\%$, and the overlap between PTB-XL (train) and Ribeiro is $46.36\%$. These results are depicted in 
Figure~\ref{fig:ptb_distributions_counts} and 
Table~\ref{tab:PTB_distances}. As per the results of this experiment, it can be concluded that PTB-XL training and test sets are considered ID with each other while being considered OOD with Chapman and Ribeiro datasets.

\begin{table}[t]
    \caption{Quantitative comparison of the distribution overlap between PTB-XL train and test sets (ID) and PTB-XL with Chapman and Ribeiro datasets (OOD).}
    \centering
    \begin{tabular}{c c } 
    \toprule
         \textbf{Datasets} & \textbf{Overlap with PTB-XL Train Set} \\ \hline\hline
         PTB-XL Test Set &  $83.52\%$\\
         Chapman & $63.65\%$ \\
         Ribeiro & $46.36\%$ \\ 
         \bottomrule
    \end{tabular}
    
    \label{tab:PTB_distances}
\end{table}

\begin{figure}[!t]
    \centerline{\includegraphics[width=.95\columnwidth]{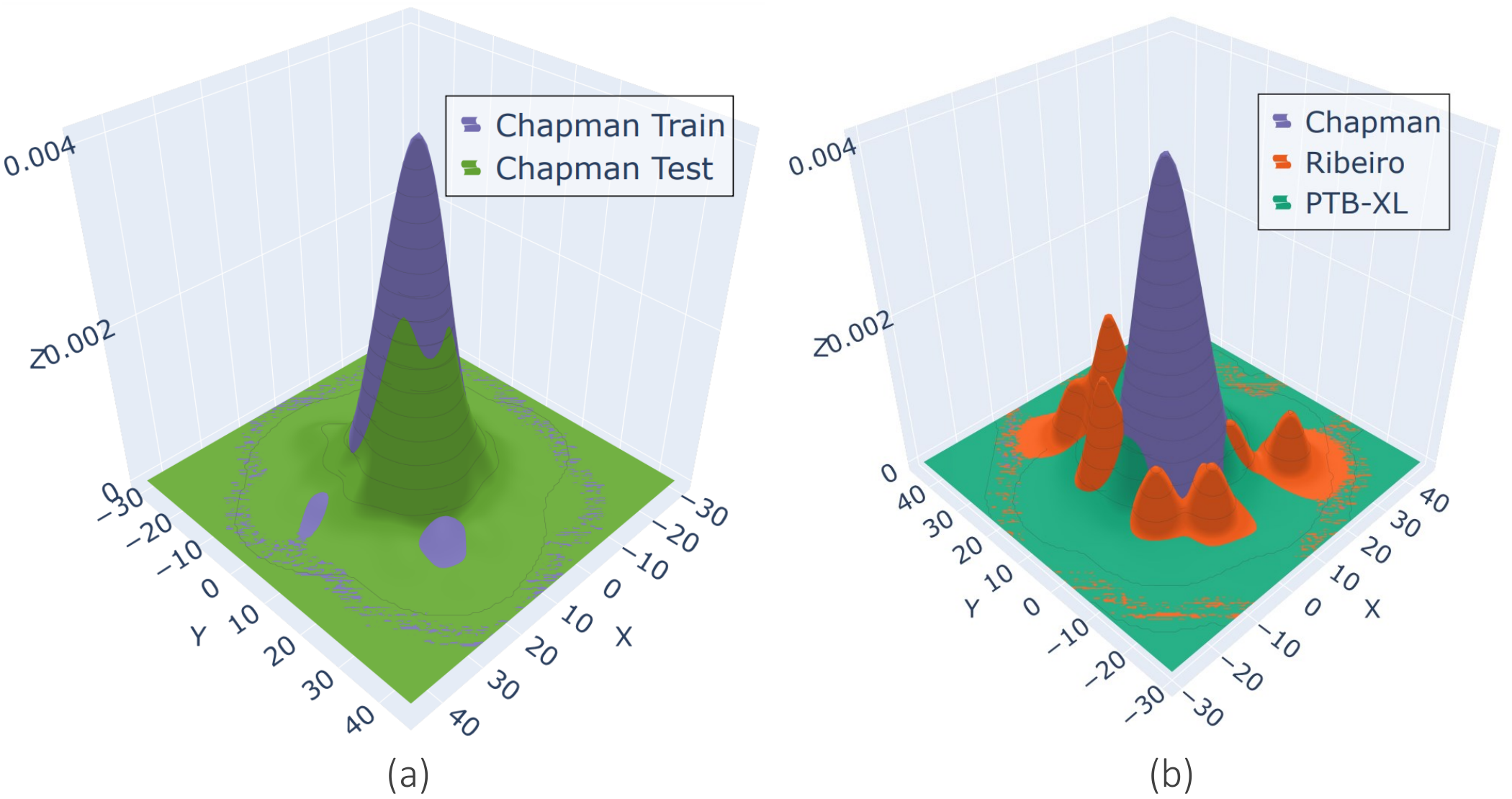}}
\caption{Comparing ID (Chapman train and test sets) against OOD (Chapman train set, PTB-XL, and Ribeiro) distribution overlaps through fitted KDE on the representations (left: ID, right: OOD).}
\label{fig:chapman_distributions_count}
\end{figure}

Next, we apply the same procedure to Chapman train and test sets. Figure~\ref{fig:chapman_distributions} displays the results of this analysis on the Chapman dataset.
Our results, which are shown in Table~\ref{tab:chapman_distributions_chount}, show that the overlap between Chapman train and test sets is estimated as $82.52\%$. We also perform the procedure using PTB-XL and Ribeiro datasets to find the overlap with Chapman train set. The overlap between the train set of Chapman dataset and PTB-XL dataset is calculated as $67.73\%$, while the overlap with Ribeiro dataset is estimated to be $60.29\%$.
These results are depicted in 
Figure~\ref{fig:chapman_distributions_count}.
These results confirm that PTB-XL and Ribeiro datasets can be considered OOD with respect to Chapman train set since their overlap is significantly lower than that of Chapman train and test sets.

\begin{table}[t]
    \centering
    \caption{Quantitative comparison of the distribution overlap between Chapman train and test sets (ID) and Chapman with PTB-XL and Ribeiro datasets (OOD).}
    \begin{tabular}{cc} 
    \toprule
         \textbf{Datasets} & \textbf{Overlap with Chapman Train Set} \\ \hline\hline
         Chapman Test Set &  $82.52\%$\\
         PTB-XL & $67.73\%$ \\
         Ribeiro & $60.29\%$ \\ 
         \bottomrule
    \end{tabular}
    \label{tab:chapman_distributions_chount}
\end{table}

\begin{table*}[t]
    \centering
    \caption{Performance of SimCLR, SwAV, and BYOL in arrhythmia classification from ECG signals on PTB-XL, Chapman, and Ribeiro Datasets. The performances are presented in the following format: F1 (\%) / Precision (\%) / Recall (\%). The reported F1 values are the best scores across all the tested augmentation settings (described in Section III-D). \textcolor{black}{The dataset names in the second row represent those used for pre-training, while the dataset names in the first column are the testing datasets.}}
    \setlength
    \tabcolsep{2pt}
    \resizebox{\textwidth}{!}{
    \begin{tabular}
    {c c|c|c|c|c|c|c}

     \toprule
  & 
 \multicolumn{2}{c}{\textbf{\textbf{SimCLR}}} & \multicolumn{2}{c}{\textbf{BYOL}} & \multicolumn{2}{c}{\textbf{SwAV}} & \textbf{Supervised}\\ 
    \cmidrule(lr){2-3} \cmidrule(lr){4-5} \cmidrule(lr){6-7} 
    \textit{Pre-training} & \textit{Chapman} & \textit{PTB-XL} & \textit{Chapman} & \textit{PTB-XL} & \textit{Chapman} & \textit{PTB-XL} & \\
\midrule

\textbf{PTB-XL} & $84.38$ / $90.92$ / $79.13$ &  $84.65 $ / $91.92$ / $78.50$ & $84.50$ / $90.60$ / $78.82$ & $85.09 $ / $91.20$ / $77.95$ & $\textbf{85.53 / 92.11 / 78.32}$ & $85.43 $ / $90.49$ / $77.44$ & $80.24 $ / $88.20$ / $74.53$ \\

Aug. & Baseline Wander & Baseline Wander & Masking & Masking & Dynamic Time Warp & Masking & - \\ 

param. & 1 & 1 & [0\%-10\%] & [40\%-50\%] & [3, 5]& [40\%-50\%] & - \\ \midrule

\textbf{Chapman} & $71.95$ / $80.53$ / $65.97$ & $71.89$ / $80.25$ / $65.27$ & $70.69$ / $79.74$ / $65.16$ & $69.54$ / $79.09$ / $64.27$ & $\textbf{72.19 / 81.86 / 66.95}$ & $70.47$ / $78.59$ / $65.75$ & $70.96$ / $78.95$ / $64.59$\\

Aug. & Dynamic Time Warp & Channel Resize &  EMG Noise & Channel Resize & Dynamic Time Warp &  EMG Noise & - \\ 
param. & [3, 10] & [0.33-3] & 0.5 & [40\%-50\%] & [1, 10]& 0.01 & - \\ \midrule

\textbf{Ribeiro} & $98.5$ / $99.63$ / $98.83$ & $97.93$ / $99.38$ / $96.97$ & $98.86$ / $100$ / $97.49$ & $97.39$ / $99.35$ / $96.64$ & $\textbf{99.10 / 100 / 98.86}$ & $97.97$ / $99.72$ / $96.68$ & $84.84$ / $90.29$ / $80.67$ \\
Aug. & Masking & Dynamic Time Warp & Gaussian Noise & Dynamic Time Warp & Masking & Gaussian Noise & - \\ 
param. & [40\%-50\%] & [1, 10] & 0.01 & [3, 5] & [0\%-50\%] & 0.01 & - \\ 

\bottomrule
    \end{tabular}
    }
    \label{tab:best_F1_scores}
\end{table*}

\subsection{SSL Method Performance}
In this section, we conduct an extensive experiment to obtain a comprehensive evaluation on the performance of SSL methods for ECG-based arrhythmia detection. Based on the results from our data distribution analysis (Section~\ref{app:data_distribution_analysis}) 
we use cross-dataset training and testing with 
Chapman and PTB-XL (train sets) used for training, while PTB-XL, Chapman, and Ribeiro datasets are used for testing. To follow a fair comparison between different methods and evaluate only the impact of the self-supervised aspect of the models, we use a standard encoder (xResNet1d50) throughout all of these studies. 
Our findings indicate that each of the SSL methods exhibit unique strengths and weaknesses, providing valuable insights into the performance of these approaches for ECG-based arrhythmia detection in ID and OOD settings. 

Table~\ref{tab:best_F1_scores} presents the results for the \textit{highest}  f1 scores obtained across all augmentation parameters and settings described in Sections III-C and III-D, respectively. In addition to individual augmentations, we also explored various combinations of augmentations. While combining augmentations were consistently among the strong performers, they did not show a clear advantage over the best \textit{individual} augmentations, and therefore were not shown in the table.
As evident by the results, across all datasets, the performance of the SSL methods outperforms that of fully supervised training. These findings suggest that SSL methods have significant potential for learning ECG representations and can be effectively leveraged for arrhythmia classification after finetuning.

Moreover, we observe that the SwAV method outperforms SimCLR and BYOL. For instance, when the model is pre-trained on Chapman and tested on the same dataset (ID), SwAV achieves an  f1 score of $72.19\%$, which is higher than that of SimCLR ($71.95\%$) and BYOL ($70.69\%$) as shown in Table~\ref{tab:best_F1_scores}. Similarly, when the model is pre-trained on PTB-XL dataset and tested on Chapman dataset (OOD), SwAV achieves a higher  f1 score ($70.47\%$) than SimCLR ($71.89\%$) and BYOL ($96.54\%$). Furthermore, when pre-trained on PTB-XL and tested on the same dataset (ID), SwAV achieves an  f1 score of $85.43\%$, which is higher than SimCLR ($84.65\%$) and BYOL ($85.09\%$). Finally, when pre-trained on Chapman dataset and tested on PTB-XL dataset (OOD), SwAV achieves an  f1 score of $85.53\%$, while SimCLR and BYOL achieve  f1 scores of $84.38\%$ and $84.50\%$, respectively. 
Additionally, when pre-training the model using Chapman dataset and testing on Ribeiro dataset (OOD), the results show that SwAV obtains an  f1 score of $99.10\%$, again outperforming SimCLR ($98.54\%$) and BYOL ($98.86\%$).
When pre-training the models on PTB-XL dataset and testing them on Ribeiro (OOD), SwAV obtains an  f1 score of $97.97\%$ outperforming the other two methods, SimCLR ($97.93\%$) and BYOL ($97.39\%$). 
The results above confirm that SwAV is able to learn more effective representations for ECG-based arrhythmia detection, followed by SimCLR and BYOL. Prior work~\cite{mehari2022self} has also shown the superiority of SimCLR over BYOL in the context of ECG.

While Table~\ref{tab:best_F1_scores} presented the results using the `best' settings, Table~\ref{tab:avg_F1_scores_2} presents the performance metrics averaged over all augmentation settings (described in Section III-D).
Results again show that the SSL methods outperform the fully supervised model. These results also confirm that SwAV exhibits the best performance in terms of  f1 scores among the three SSL methods.
We hypothesize that the clustering-based approach in SwAV might have allowed the model to learn more robust and diverse representations that capture the essential and generalizable features of the ECG signals, leading to better performance. Clustering can be especially effective in cases where there are different underlying structures in the data, as it can help to identify this structure and improve the quality of the learned representations~\cite{vagner2011clustering, yeh2012analyzing, he2019unsupervised}.

\begin{table*}[t]
    \centering
    \caption{Performance of SimCLR, SwAV, and BYOL in arrhythmia classification from ECG signals on PTB-XL, Chapman, and Ribeiro datasets. The performances are presented in the following format: F1 (\%) / Precision (\%) / Recall (\%).
    The reported F1 values are the average scores obtained across all augmentation settings (described in Section III-D). \textcolor{black}{The dataset names in the second row represent those used for pre-training, while the dataset names in the first column are the testing datasets.}}
    \setlength
    \tabcolsep{2pt}

    \resizebox{\textwidth}{!}{
    
    \begin{tabular}{c c|c|c|c|c|c|c}
    \toprule
  & 
 \multicolumn{2}{c}{\textbf{\textbf{SimCLR}}} & \multicolumn{2}{c}{\textbf{BYOL}} & \multicolumn{2}{c}{\textbf{SwAV}} & \textbf{Supervised} \\ 
    \cmidrule(lr){2-3} \cmidrule(lr){4-5} \cmidrule(lr){6-7}
\textit{Pre-training} & \textit{Chapman} & \textit{PTB-XL} & \textit{Chapman} & \textit{PTB-XL} & \textit{Chapman} & \textit{PTB-XL} & \\
\midrule
\textbf{PTB-XL} & $83.44$ / $89.63$ / $79.27$ & $83.54$ / $89.37$ / $79.84$ & $83.11$ / $88.08$ / $79.37$ & $82.91$ / $88.67$ / $78.93$ & $\textbf{84.22 / 91.03 / 78.27}$ & $83.95$ / $89.84$ / $79.62$ & $75.22$ / $79.25$ / $73.85$\\

\textbf{Chapman} & $66.99$ / $77.52$ / $59.63$ & $65.69$ / $76.53$ / $58.36$ & $67.15$ / $78.26$ / $61.62$ & $65.36$ / $76.38$ / $58.85$ & $\textbf{67.99 / 77.08 / 60.70}$ & $66.38$ / $77.27$ / $59.26$ & $63.25$ / $69.56$ / $60.40$\\

\textbf{Ribeiro} & $96.40$ / $98.63$ / $96.25$ & $95.13$ / $98.30$ / $94.37$ & $96.89 $ / $99.26$ / $96.80$ & $94.12 $ / $98.04$ / $92.64$ & $\textbf{97.28 / 99.97 / 97.07}$ & $94.79 $ / $98.57$ / $92.08$ & $79.93 $ / $83.34$ / $78.57$\\
\bottomrule
    \end{tabular}
    }
    \label{tab:avg_F1_scores_2}
\end{table*}

\begin{table}[t]
    \centering
    \caption{A comparison of the SSL results in comparison to a number of baselines tested on PTB-XL dataset (on the left), and Chapman dataset (on the right). The baseline results for PTB-XL dataset have been reported in~\cite{strodthoff2020deep}. }
    \begin{tabular}{lc|lcc}
    \toprule
         \multicolumn{2}{c}{\textbf{PTB-XL}} &  \multicolumn{3}{c}{\textbf{Chapman}} \\
         \cmidrule(lr){1-2} \cmidrule(lr){3-5}
        \textbf{Method} & \textbf{AUC} & \textbf{Method} & \textbf{Ref.} & \textbf{AUC}  \\ \midrule
        LSTM & $92.7\%$ & CPC & \cite{falcon2020framework} & $84.4\%$ \\  
        Inception1d & $93.1\% $ & SSLECG & \cite{sarkar2020self} & $52.6\%$ \\  
        LSTM\_bidir & $93.2\%$ & CLOCS & \cite{kiyasseh2021clocs} & $90.6\%$ \\  
        Resnet1d\_wang & $93.6\% $ & w/o Inter & \cite{lan2022intra} & $76.4\%$ \\  
        FCN\_awng & $92.6\% $ & w/o Intra & \cite{lan2022intra} & $92.1\%$ \\  
        Wavelet+NN & $85.5\%$ & ISL & \cite{lan2022intra} & $96.5\%$ \\  
        xResnet1d101 & $93.7\%$ & \textit{SimCLR} & - & $93.5\%$ \\  
        Ensemble & $93.9\%$ & \textit{BYOL} & - & $94.9\%$ \\  
        \textit{SimCLR} & $93.6\% $ & \textit{SwAV} & - & $\textbf{96.7\%}$ \\  
        \textit{BYOL} & $93.6\%$ & & & \\  
        \textit{SwAV} & $\textbf{94.0\%}$& & & \\ 
        \bottomrule
    \end{tabular}
    \label{tab:compare_to_lit_ptb}
\end{table}

In Table~\ref{tab:compare_to_lit_ptb}, we present a comparison between the SSL methods 
and popular models for PTB-XL and Chapman datasets.
The models presented for PTB-XL dataset are reported in~\cite{strodthoff2020deep}, and for Chapman dataset, the models are from the literature. To maintain consistency with other methods and for a fair comparison, we use the 4-class format of Chapman dataset in this part of our study.
For both of the datasets, the SwAV method shows the highest F1 among all the methods. 

\textcolor{black}{As part of our methodology, we also conduct linear evaluations to assess the performance of the SSL models. This involves freezing the weights of the model after pre-training and adding a new fully connected layer. We then train this layer using labeled data without making any modifications to the pre-existing weights. The results are presented in Table~\ref{tab:linear_evaluation_results}. As expected, linear evaluation achieves lower scores than the fine-tuning approach.} However, linear evaluation still provides valuable insights into the effectiveness of the SSL models in ID and OOD settings. It allows us to measure the extent to which the models have learned useful and generalized representations during pre-training. The results (presented in Table~\ref{tab:linear_evaluation_results}) show that the model has performed reasonably well, even though the encoder is kept frozen, and this in turn shows that the model has learned meaningful and effective representations during the self-supervised pre-training.

\noindent \textbf{Discussion.} 
First, by comparing Tables\ref{tab:best_F1_scores} and \ref{tab:linear_evaluation_results}, we expectedly notice that fine-tuning the model on the target dataset will boost performance considerably for both ID and OOD schemes. However, we make an interesting and unexpected observation about the nature of ID and OOD ECG-based arrhythmia detection with SSL. By considering the results of Tables \ref{tab:best_F1_scores} and \ref{tab:linear_evaluation_results}, and comparing ID to OOD results within each table, we notice that in almost every scenario, the performance of the models are the same between OOD and ID setting. While this observation would be more expected with the fine-tuning setup, we even notice this trend with linear evaluation. This finding indicates that irrespective of dataset, equipment, and population of patients, SSL methods are able to effectively learn representations that can be used to accurately detect different types of arrhythmia.
For instance, in the case of training SwAV on Chapman dataset and testing on PTB-XL, despite the relatively OOD nature of these two datasets, we still witnessed comparable performance levels as when pre-training using PTB-XL dataset and testing on the same dataset (see Table~\ref{tab:best_F1_scores} and Table~\ref{tab:linear_evaluation_results} for the reported results). This outcome suggests that the representations learned by SSL models can transfer effectively across diverse datasets and domains, regardless of the distribution shifts present in the data, presenting important implications for practical applications.

\begin{figure}[!t]
    \centerline{\includegraphics[width=0.80\columnwidth]{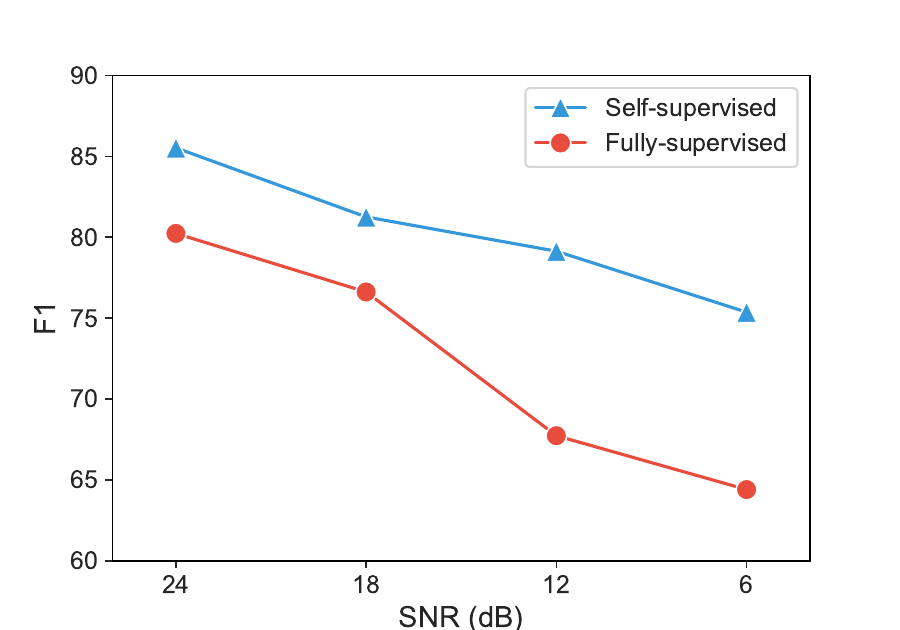}}
    \caption{F1 scores versus SNR for the best-performing SSL method, SwAV, and the fully-supervised baseline, trained on PTB-XL dataset. This figure illustrates the robustness of SSL against noise.}
    \label{f1_to_snr}
\end{figure}

To investigate the robustness of SSL models against noise, we introduce different amounts of synthetic $60$ Hz (power line) interference to PTB-XL test set, as a common type of noise often present in ECG datasets \cite{kher2019signal, bahaz2018efficient}. We evaluate the performance of the highest-performing SSL method, SWaV, 
trained on PTB-XL dataset with dynamic time warp augmentation, alongside the fully-supervised model. The F1 score plotted against the signal-to-noise ratio (SNR) in Figure~\ref{f1_to_snr}, demonstrates that the addition of noise has a smaller impact on the SSL model's performance compared to the fully-supervised model as a lower degradation is observed.

\begin{table}[t]
    \centering
    \caption{Linear evaluation on Chapman, PTB-XL and Ribeiro datasets. \textcolor{black}{The dataset names in the second row represent those used for pre-training, while the dataset names in the first column are the testing datasets.}}
    \setlength
    \tabcolsep{3pt}
    \scriptsize
    \begin{tabular}{ccccccc}
    \toprule
  & 
 \multicolumn{2}{c}{\textbf{\textbf{SimCLR}}} & \multicolumn{2}{c}{\textbf{BYOL}} & \multicolumn{2}{c}{\textbf{SwAV}}\\ 
    \cmidrule(lr){2-3} \cmidrule(lr){4-5} \cmidrule(lr){6-7} 
    \textit{Pre-training} & \textit{Chapman} & \textit{PTB-XL} & \textit{Chapman} & \textit{PTB-XL} & \textit{Chapman} & \textit{PTB-XL} \\
\midrule
\textbf{PTB-XL} & $76.03\%$ & $76.57\%$ & $75.97\%$ & $76.67\%$ & $\textbf{79.61\%}$ & $79.60\%$ \\

\textbf{Chapman} & $49.79\%$ & $48.55\%$ & $48.30\%$ & $47.53\%$ & $\textbf{50.37\%}$ & $49.85\%$ \\

\textbf{Ribeiro} & $90.02\%$ & $87.89\%$ & $86.19\%$ & $84.32\%$ & $\textbf{93.37\%}$& $86.96\%$ \\
\bottomrule
    \end{tabular}
    
    \label{tab:linear_evaluation_results}
\end{table}

\subsection{Per-Disease Analysis}
Cardiovascular diseases encompass a wide range of conditions, some of which can be rare and life-threatening. In this section, we aim to evaluate the performance of SSL in the classification of each cardiovascular disease separately. 
To this end, we first present the per-class f1 scores of the best performing models for PTB-XL dataset in Table~\ref{tab:best_model_perclass_ptb}, \textcolor{black}{which examines the performance of the model on each disease class separately to allow a detailed assessment of how well the model predicts each individual class}. Overall, the best performance of the models on the Normal class of PTB-XL dataset is achieved when SwAV is pre-trained using Chapman dataset, with an  f1 score of $88.14\%$. The CD and MI  classes also have relatively strong performances, with  f1 scores of $80.85\%$ and $79.77\%$, respectively, when pre-trained on PTB-XL dataset. However, identifying instances from the HYP and STTC classes prove more challenging for the models, with  f1 scores of $71.96\%$ and $77.42\%$, respectively. This could indicate that features corresponding to some diseases are slightly more challenging to learn compared to others.

Table~\ref{tab:best_model_perclass_ptb} demonstrates that the SwAV model pre-trained on PTB-XL outperforms other models for each class of PTB-XL dataset, with the exception of the normal class, where pre-training on Chapman yields the best result.

\begin{table}[t]
    \centering
    \caption{Per-class 
    results for PTB-XL dataset. \textcolor{black}{The dataset names in the second row represent those used for pre-training, while the dataset names in the first column are the testing datasets.}}
    \setlength
    \tabcolsep{3pt}
    \scriptsize
    \begin{tabular}{lcccccc}
    \toprule
& \multicolumn{2}{c}{\textbf{\textbf{SimCLR}}} & \multicolumn{2}{c}{\textbf{BYOL}} & \multicolumn{2}{c}{\textbf{SwAV}} \\ 
    \cmidrule(lr){2-3} \cmidrule(lr){4-5} \cmidrule(lr){6-7}
\textit{Pre-training} & \textit{Chapman} & \textit{PTB-XL} & \textit{Chapman} & \textit{PTB-XL} & \textit{Chapman} & \textit{PTB-XL} \\
\midrule
\textbf{MI} & $78.05\%$ & $78.15\%$ & $77.45\%$ & $77.66\%$ & $78.96\%$ & $\textbf{79.77\%}$ \\
\textbf{HYP} & $67.82\%$ & $67.34\%$ & $66.08\%$ & $64.44\%$ & $70.07\%$ & $\textbf{71.96\%}$ \\
\textbf{Normal} & $87.81\%$ & $87.67\%$	& $87.60\%$ & $87.10\%$ & $\textbf{88.14\%}$ & $88.06\%$ \\
\textbf{STTC} & $75.78\%$ & $75.22\%$ & $74.33\%$ & $74.59\%$ & $76.62\%$ & $\textbf{77.42\%}$ \\
\textbf{CD} & $79.41\%$ & $80.15\%$ & $79.42\%$ & $79.59\%$ & $80.84\%$ & $\textbf{80.85\%}$ \\
\bottomrule
    \end{tabular}
    \label{tab:best_model_perclass_ptb}
\end{table}

To evaluate the performance of the models on Chapman dataset, we use the 11-class configuration as described in Section~\ref{app:dataset}. Although the classes are not evenly distributed, we evaluate each class separately without making any adjustments to the dataset. Per-class f1 scores are presented in Table~\ref{tab:best_f1_model_chapman}. 

\begin{table}[t]
    \centering
    \caption{Per-class results for Chapman dataset.}
    \setlength
    \tabcolsep{3pt}
    \scriptsize
    \begin{tabular}{lcccccc}\toprule
  & 
 \multicolumn{2}{c}{\textbf{\textbf{SimCLR}}} & \multicolumn{2}{c}{\textbf{BYOL}} & \multicolumn{2}{c}{\textbf{SwAV}} \\ 
    \cmidrule(lr){2-3} \cmidrule(lr){4-5} \cmidrule(lr){6-7}
\textit{Pre-training} & \textit{Chapman} & \textit{PTB-XL} & \textit{Chapman} & \textit{PTB-XL} & \textit{Chapman} & \textit{PTB-XL} \\\midrule

\textbf{AF} &  $55.58\%$ & $56.59\%$ & $\textbf{58.63\%}$ & $54.31\%$ & $57.26\%$ & $57.08\%$\\
 
\textbf{AFIB} &  $90.37\%$ & $\textbf{90.86\%}$ & $90.65\%$ & $89.81\%$ & $89.14\%$ & $88.98\%$\\
 
\textbf{AT} & $9.09\%$ & $8.99\%$ & $6.74\%$ & $6.90\%$ & $\textbf{17.39\%}$ & $8.89\%$ \\
 
\textbf{AVNRT} & $0.00\%$ & $0.00\%$ & $0.00\%$ & $0.00\%$ & $0.00\%$ & $0.00\%$\\
 
\textbf{AVRT} & $0.00\%$ & $0.00\%$ & $0.00\%$ & $0.00\%$ & $0.00\%$ & $0.00\%$\\
 
\textbf{SI} & $\textbf{31.49\%}$ & $29.94\%$ & $26.87\%$ & $30.81\%$ & $29.67\%$ & $30.32\%$\\
 
\textbf{SAAWR} & $0.00\%$ & $0.00\%$ & $0.00\%$ & $0.00\%$ & $0.00\%$ & $0.00\%$\\
 
\textbf{SB} & $96.66\%$ & $96.36\%$ & $\textbf{96.72\%}$ & $96.56\%$ & $96.45\%$ & $96.35\%$\\
 
\textbf{SR} & $86.31\%$ & $\textbf{96.09\%}$ & $86.34\%$ & $95.91\%$ & $86.22\%$ & $85.67\%$\\
 
\textbf{ST} & $94.47\%$ & $94.25\%$ & $94.42\%$ & $94.72\%$ & $\textbf{94.92\%}$ & $94.51\%$\\
 
\textbf{SVT} & $85.68\%$ & $84.22\%$ & $85.64\%$ & $85.29\%$ & $\textbf{86.60\%}$ & $85.82\%$\\
\bottomrule
    \end{tabular}
    
    \label{tab:best_f1_model_chapman}
\end{table}

\begin{figure}[!t]
    \centerline{\includegraphics[width=0.88\columnwidth]{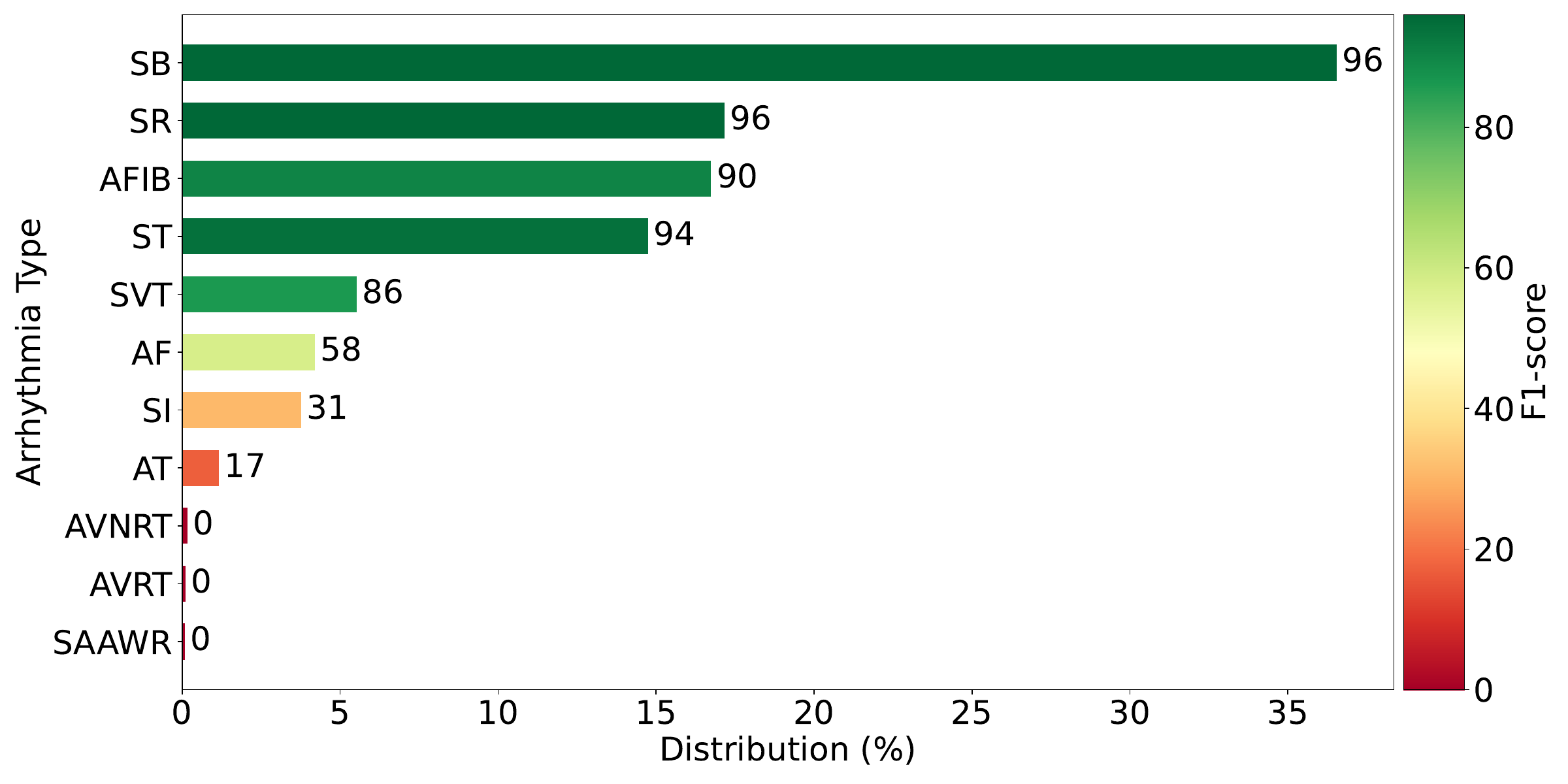}}
    \caption{
    The figure depicts the distribution of each class in Chapman dataset alongside the corresponding F1 scores, providing an overview of the relationship between class distribution and model performance.}
    \label{fig:best_F1_chapman_perclass}
\end{figure}

The number of samples in each class of Chapman dataset varies significantly, and there is a correlation between the number of samples and the model's classification performance, shown in Figure~\ref{fig:best_F1_chapman_perclass}. SB disease is the most populated class, with $36.53\%$ of the total samples, achieving the best result with an  f1 score of $96.72\%$. The next most populated class is SR, accounting for $17.15\%$ of the total samples, and achieves an  f1 score of $96.09\%$. For AFIB, ST, and SVT, the results are reasonable, but there is a drop in performance for classes with fewer samples. For instance, AF, which accounts for $4.18\%$ of the total samples, shows an  f1 score of $58.63\%$, representing a $25\%$ drop from the previous class. SI and AT, which account for $3.75\%$ and $1.14\%$ of the dataset respectively, achieve  f1 scores of $31.49\%$ and $17.39\%$. However, for AVNRT, AVRT, and SAAWR, which have less than $1\%$ distribution in the dataset, classification performances are poor.

Based on the analysis of the  f1 scores presented in Table~\ref{tab:best_f1_model_chapman}, it appears that pre-training the models on Chapman dataset has yielded superior results for the majority of the disease classes compared to pre-training on PTB-XL dataset, with the exception of AFIB and SR. 

Finally, we present the  f1 scores achieved for each disease class using the best performing model for Ribeiro dataset in Table~\ref{tab:ribeiro_perclass_F1}.
The results indicate that the BYOL method pre-trained on Chapman dataset performs better than the other cases in classifying each disease class of Ribeiro dataset, with the exception of the AFIB disease, where SimCLR pre-trained on Chapman dataset achieved the best result. This may be due to the fact that the diseases in Chapman dataset have more similarities to those in Ribeiro dataset, as described in Section~\ref{app:dataset}. Specifically, the main class of all the diseases in Ribeiro dataset belong to the CD category, which is well-represented in Chapman dataset, but less so in PTB-XL.
Considering the highest  f1 scores in Table~\ref{tab:ribeiro_perclass_F1}, we can see that despite the low number of samples used for fine-tuning, all classes are classified well, except for AFIB, which appears to be classified less accurately than the other classes.

\begin{table}[t]
    \centering
    \caption{Per-class results for Ribeiro dataset.}
    \setlength
    \tabcolsep{3pt}
    \scriptsize
    \begin{tabular}{ccccccc}\toprule
  & 
 \multicolumn{2}{c}{\textbf{\textbf{SimCLR}}} & \multicolumn{2}{c}{\textbf{BYOL}} & \multicolumn{2}{c}{\textbf{SwAV}} \\ 
    \cmidrule(lr){2-3} \cmidrule(lr){4-5} \cmidrule(lr){6-7}
\textit{Pre-training} & \textit{Chapman} & \textit{PTB-XL} & \textit{Chapman} & \textit{PTB-XL} & \textit{Chapman} & \textit{PTB-XL} \\
\hline
\textbf{1-AVB} &  $93.33\%$ & $81.08\%$ & $83.33\%$ & $82.35\%$ & $93.88\%$ & $95.24\%$\\
 
\textbf{RBBB} &  $100.00\%$ & $97.87\%$ & $97.96\%$ & $100.00\%$ & $100.00\%$ & $97.67\%$\\
 
\textbf{LBBB} & $100.00\%$ & $100.00\%$ & $100.00\%$ & $100.00\%$ & $100.00\%$ & $100.00\%$\\
 
\textbf{SBRAD} & $100.00\%$ & $100.00\%$ & $100.00\%$ & $100.00\%$ & $95.24\%$ & $100.00\%$\\
 
\textbf{AFIB} & $61.54\%$ & $66.67\%$ & $71.43\%$ & $75.00\%$ & $87.50\%$ & $60.00\%$\\
 
\textbf{STACH} & $100.00\%$ & $100.00\%$ & $100.00\%$ & $100.00\%$ & $97.96\%$ & $100.00\%$\\
\bottomrule
    \end{tabular}
    
    \label{tab:ribeiro_perclass_F1}
\end{table}

\section{Conclusion and Future Work}
In this paper we analyzed ECG-based arrhythmia detection using SSL methods, in ID and OOD settings. To this end, we studied the distributions of three popular ECG-based arrhythmia datasets to identify and quantify the their relative distributions and to determine whether they can be considered ID or OOD. Next, we implement three popular SSL methods, SimCLR, BYOL, and SwAV for ECG-based arrhythmia detection. 
In analyzing the augmentation parameters we found that in general, certain augmentation techniques such as mix of augmentations, time warping, and masking, improve the learning of ECG representations and result in better generalization, while negation augmentation showed poor performance. 
In comparing different SSL techniques with state-of-the-art, we observed highly competitive performances in both ID and OOD settings, while among the SSL methods, SwAV outperformed the others. Another interesting finding of our work is that SSL techniques can detect arrhythmia in OOD settings with competitive performance to that of ID settings.
Lastly, we conducted a per-class analysis for each dataset to understand the performance of the SSL methods toward the detection of different diseases and found that false negative errors are more common than false positives.

A number of future directions can be considered to improve upon the work presented in this paper. 
To broaden our research, we can consider delving into finer step sizes and wider parameter ranges for each augmentation to determine the ideal augmentation parameters. Furthermore, we could examine more sophisticated augmentations, such as those implemented in the frequency domain.
In this study, we explored each augmentation individually, and the mix of all of the augmentations together. Nonetheless, different combinations of augmentations may exhibit various behaviours, which could be explored through forward/backward selection methodologies in the future. 
In terms of the SSL methods considered, more methods could be explored, and the number and type of architectures and backbone encoders could also be expanded.
Additionally, we aim to conduct a more in-depth investigation into the influence of each individual SSL method on data distribution by incorporating additional visualizations, to provide a deeper understanding of how these methods shape meaningful features within the ECG data.
Lastly, using a combination of different datasets for pre-training the SSL models may result in learning more generalizable ECG representations, which could further improve the performance in OOD settings.

\section*{Acknowledgement} This project was funded in part by Natural Sciences and Engineering Research Council of Canada.

\footnotesize
\bibliographystyle{ieeetr} 
\bibliography{IEEEfull}
\end{document}